%% file: main.tex
\documentclass[letterpaper, 10 pt, journal, twoside]{IEEEtran}
\usepackage{amsmath,amsfonts}
\usepackage{algorithmic}
\usepackage{algorithm}
\usepackage{array}
\usepackage{textcomp}
\usepackage{stfloats}
\usepackage{url}
\usepackage{verbatim}
\usepackage{graphicx}
\usepackage{cite}

\usepackage[utf8]{inputenc} 
\usepackage[T1]{fontenc}    
\usepackage{hyperref}       
\usepackage{url}            
\usepackage{booktabs}       
\usepackage{nicefrac}       
\usepackage{microtype}      
\usepackage{xcolor}         
\usepackage{colortbl}
\usepackage[english]{babel}
\usepackage{amsthm}
\newtheorem*{remark}{Remark}
\usepackage{tikz}
\usepackage{comment}
\usepackage[bb=boondox]{mathalfa}
\usepackage{multirow}
\usepackage{wrapfig}
\usepackage{rotating}
\usepackage{subfigure}
\usepackage{subcaption}

\usepackage{marvosym}
\usepackage{makecell}
\usepackage{pifont}
\usepackage{float}
\usepackage{rotating} 
\usepackage[capitalize,noabbrev]{cleveref}
\crefname{algorithm}{Alg.}{Algs.}
\Crefname{algocf}{Algorithm}{Algorithms}
\crefname{section}{Sec.}{Secs.}
\crefname{table}{Tab.}{Tabs.}
\crefname{figure}{Fig.}{Fig.}
\Crefname{equation}{Eq.}{Eqs.}
\crefname{appendix}{Appx.}{Appx.}
\Crefname{appendix}{Appendix}{Appendix}

\newcommand{\cmark}{\textcolor{green}{\ding{51}}}%
\newcommand{\xmark}{\textcolor{red}{\ding{55}}}%

\definecolor{customgreen}{HTML}{ccccff}
\definecolor{customblue}{HTML}{ffcccc}
\definecolor{customyellow}{HTML}{BDD7EE}

\begin{document}

\title{\bf
FlightBench: Benchmarking Learning-based Methods for Ego-vision-based Quadrotors Navigation}

\author{Shu-Ang Yu$^{1*}$, Chao Yu$^{1*{\dag}}$, Feng Gao$^{1*}$, Yi Wu$^{1,2}$, and Yu Wang$^{1{\dag}}$%
\thanks{This research was supported by National Natural Science Foundation of China (No.62406159, 62325405), Postdoctoral Fellowship Program of CPSF under Grant Number GZC20240830, China Postdoctoral Science Special Foundation 2024T170496.} 
\thanks{* Equal contribution.}%
\thanks{{\dag} Corresponding Authors. \{yuchao, yu-wang\}@mail.tsinghua.edu.cn}%
\thanks{The authors are with $^{1}$Tsinghua University, $^{2}$Shanghai Qi Zhi Institute.}%
}


\maketitle

\input{0_abs}
\input{1_intro}
\input{2_related}

\input{4_method}
\input{5_exp}

\input{6_con}

\bibliographystyle{IEEEtran}
\bibliography{reference}

\end{document}

%% file: 0_abs.tex
\begin{abstract}
Ego-vision-based navigation in cluttered environments is crucial for mobile systems, particularly agile quadrotors. While learning-based methods have shown promise recently, head-to-head comparisons with cutting-edge optimization-based approaches are scarce, leaving open the question of where and to what extent they truly excel.
In this paper, we introduce FlightBench, the first comprehensive benchmark that implements various learning-based methods for ego-vision-based navigation and evaluates them against mainstream optimization-based baselines using a broad set of performance metrics. More importantly, we develop a suite of criteria to assess scenario difficulty and design test cases that span different levels of difficulty based on these criteria. 
Our results show that while learning-based methods excel in high-speed flight and faster inference, they struggle with challenging scenarios like sharp corners or view occlusion. Analytical experiments validate the correlation between our difficulty criteria and flight performance.Moreover, we verify the trend in flight performance within real-world environments through full-pipeline and hardware-in-the-loop experiments. We hope this benchmark and these criteria will drive future advancements in learning-based navigation for ego-vision quadrotors. Code and documentation are available at \url{https://github.com/thu-uav/FlightBench}.
\end{abstract}


%% file: 1_intro.tex
\section{Introduction}
\label{sec:intro}
Ego-vision-based navigation in cluttered environments is a fundamental capability for mobile systems and has been widely investigated~\cite{xiao2023collaborative, xiao2021learning, stachowicz2023fastrlap}. It involves navigating a robot to a goal position while avoiding collisions with obstacles, using equipped ego-vision cameras~\cite{agarwal2023legged}. Quadrotors, known for their agility and dynamism~\cite{loquercio2021learning}, present unique challenges in achieving fast and safe flight. Traditionally, hierarchical methods address this problem by decoupling it into subtasks such as mapping, planning, and control~\cite{xiao2024visionbased}, optimizing the trajectory to avoid collisions. In contrast, recent works~\cite{song2023learning,kaufmann2023champion} have demonstrated that learning-based methods can unleash the full dynamic potential of agile platforms. These methods employ neural networks to generate a sequence of waypoints~\cite{loquercio2021learning} or motion commands~\cite{song2023learning, kaufmann2023champion}. Unlike the high computational costs associated with sequentially executed subtasks, this end-to-end manner significantly reduces processing latency and enhances agility~\cite{loquercio2021learning}.

Despite the promising results of learning-based navigation methods, the lack of head-to-head comparisons with state-of-the-art optimization-based methods makes it unclear in which areas they truly outperform and to what degree. Traditional methods are often evaluated using customized scenarios and sensor configurations~\cite{ren2022bubble, zhou2019robust, song2023learning}, which complicates reproducibility and hinders fair comparisons. Moreover, the absence of a quantifiable approach for scenario difficulty further obscures the analysis of the strengths and weaknesses of current methods.

\begin{figure}[t]
    \centering
    \includegraphics[width=1.0\linewidth]{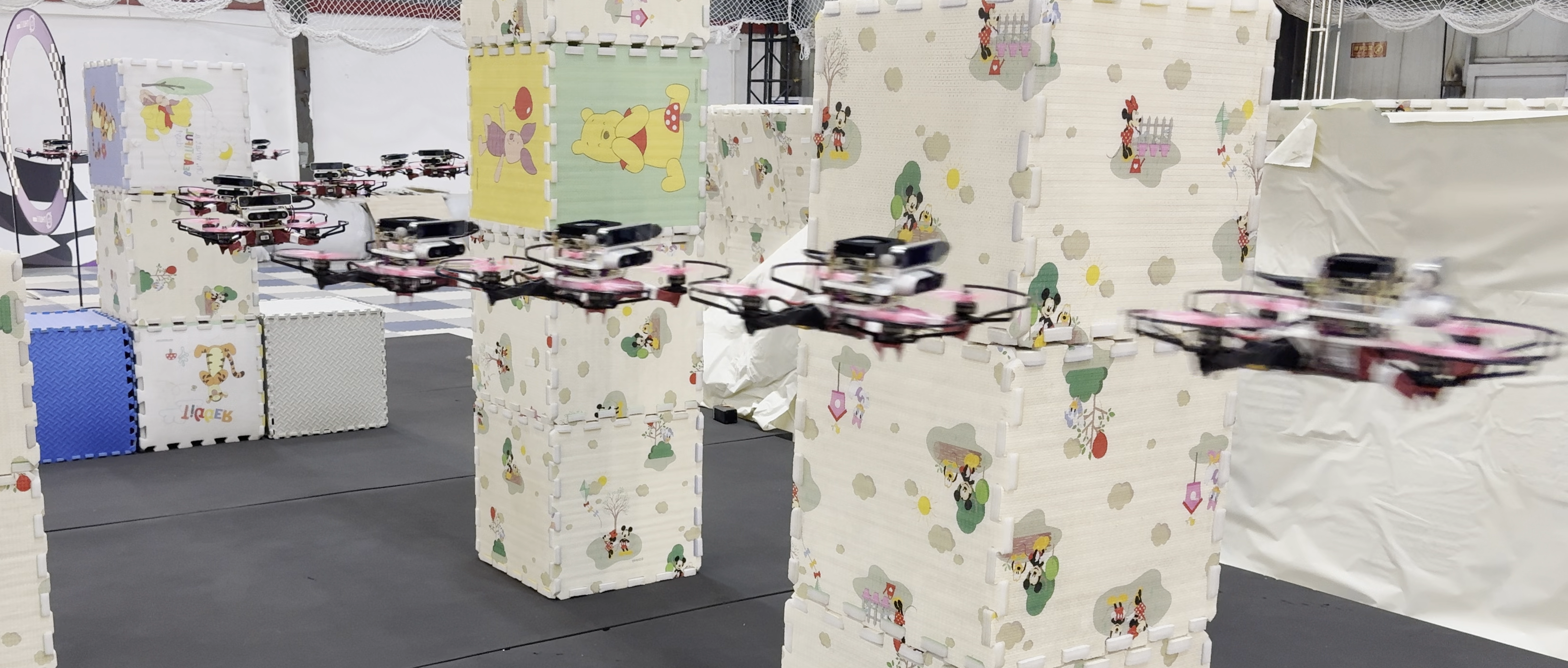}
    \vspace{-2mm}
    \caption{Real-world validation of FlightBench.}
    \label{fig:realworld}
    \vspace{-4mm}
\end{figure}
In this paper, we introduce FlightBench, a comprehensive benchmark that evaluates methods for ego-vision-based quadrotor navigation.
We initially incorporated a suite of representative navigation methods, encompassing both ego-vision and privileged ones for an in-depth comparative analysis.
Moreover, we established three criteria to measure the difficulty of scenarios, thereby creating a diverse array of tests that span a spectrum of difficulties.
Finally, we compared these methods through simulated and real-world experiments, evaluating a wide range of performance metrics to gain a deeper understanding of their specific attributes.
Our experiments indicate that learning-based methods demonstrate superior performance in high-speed flight scenarios and generally offer quicker inference times. However, they encounter difficulties in handling complex situations such as sharp turns or occluded views. 
In contrast, our findings show that traditional optimization-based methods perform well in challenging conditions, not just in success rate and flight quality, but also in computation time, particularly when they are meticulously designed.
Our analytical experiments validate the effectiveness of the proposed criteria and emphasize the importance of latency randomization for learning-based methods.
In summary, our key contributions include:

\begin{enumerate}
    \item The development of FlightBench, the first unified open-source benchmark that facilitates the head-to-head comparison of learning-based and optimization-based methods on ego-vision-based quadrotor navigation under various 3D scenarios.
    \item The proposition of tailored task difficulty and performance metrics, aiming to enable a thorough evaluation and in-depth analysis of specific attributes for different methods.
    \item Detailed experimental analyses in both simulation and real-world that demonstrate the comparative strengths and weaknesses of learning-based versus optimization-based methods, particularly in difficult scenarios.
\end{enumerate}

%% file: 2_related.tex
\section{Related Work}
\subsection{Planning methods for navigation.}
Classical navigation algorithms typically use search or sampling to explore the configuration or state space and generate a free path~\cite{kamon1997sensory, penicka2022minimum}. With optimization, a multi-objective optimization problem is often formulated to determine the optimal trajectory~\cite{paull2012sensor, ye2022efficient}. This is commonly done using gradients from local maps, such as the Artificial Potential Field (APF)~\cite{sfeir2011improved} and the Euclidean Signed Distance Field (ESDF)~\cite{zhou2019robust}. 
On the other side, the development of deep learning enables algorithms to perform navigating directly from sensory inputs such as images or lidar~\cite{xiao2024visionbased}. The policies are trained by imitating expert demonstrations~\cite{8798720} or through exploration under specific rewards~\cite{pmlr-v229-liu23a, heeg2024learning, xi2024lightweight}. Learning-based algorithms have been applied to various mobile systems, such as quadrupedal robots~\cite{agarwal2023legged}, wheeled vehicles~\cite{chaplot2020learning, stachowicz2023fastrlap}, and quadrotors~\cite{kaufmann2023champion, xing2024bootstrapping, geles2024demonstrating}. In this work, we examine representative methods for ego-vision-based navigation on quadrotors, including three learning-based approaches and three optimization-based methods, providing a comprehensive comparison between these categories. 

\begin{table}[htbp]
    \caption{A comparison of FlightBench to other open-source benchmarks for navigation.}
    \centering
    \scalebox{0.95}{
    \begin{tabular}{ccccc}
    \toprule
        Benchmark & \makecell[c]{3D Scenarios} & \makecell[c]{Classical\\Methods} & \makecell[c]{Learning\\Methods} & \makecell[c]{Sensory\\Input} \\ \midrule
        MRBP 1.0~\cite{wen2021mrpb} & \xmark & \cmark & \xmark & LiDAR  \\ 
        Bench-MR~\cite{heiden2021bench} & \xmark & \cmark & \xmark & -  \\ 
        PathBench~\cite{toma2021pathbench} & \cmark & \cmark & \cmark & -  \\ 
        Gibson Bench~\cite{xia2020interactive} & \xmark & \xmark & \cmark & Vision  \\ 
        OMPLBench~\cite{moll2015benchmarking} & \cmark & \cmark & \xmark & -  \\ 
        RLNav~\cite{xu2023benchmarking} & \xmark & \cmark & \cmark & LiDAR  \\ 
        Plannie~\cite{rocha2022plannie} & \cmark & \cmark & \cmark & - \\
        MP design~\cite{shao2024design} & \cmark & \cmark & \xmark & - \\
        \textbf{FlightBench~(Ours)} & {\cmark} & \cmark & \cmark & Vision  \\ \bottomrule
    \end{tabular}
    }
    \label{tab:related}
    \vspace{-3mm}
\end{table}

\subsection{Benchmarks for navigation.} 
\label{sec:related_benchmarks}
Several benchmarks exist for non-sensory-input navigation algorithms. OMPL~\cite{moll2015benchmarking} and Bench-MR~\cite{toma2021pathbench} primarily focus on sampling-based methods, while PathBench~\cite{toma2021pathbench} evaluates graph-based and learning-based methods. Plannie~\cite{rocha2022plannie} offers sampling-based, heuristic, and learning-based methods for quadrotors. The most closely related work is MP design~\cite{shao2024design}, which lacks vision-input baselines and does not consider the visual perception challenges in its proposed ECS metric.
For methods with sensory inputs, most benchmarks are primarily designed for 2D scenarios. MRBP1.0~\cite{wen2021mrpb} and RLNav~\cite{xu2023benchmarking} evaluate planning methods using laser-scanning data for navigation around columns and cubes. GibsonBench~\cite{xia2020interactive} features a mobile agent equipped with a camera, navigating in interactive environments. As outlined in \cref{tab:related}, there's a notable lack of a benchmark with 3D scenarios and ego-vision inputs to assess and compare both classical and learning-based navigation algorithms, a gap that FlightBench aims to fill.

%% file: 4_method.tex
\section{FlightBench}
\begin{figure*}[!t]
    \centering
    \includegraphics[width=0.95\linewidth]{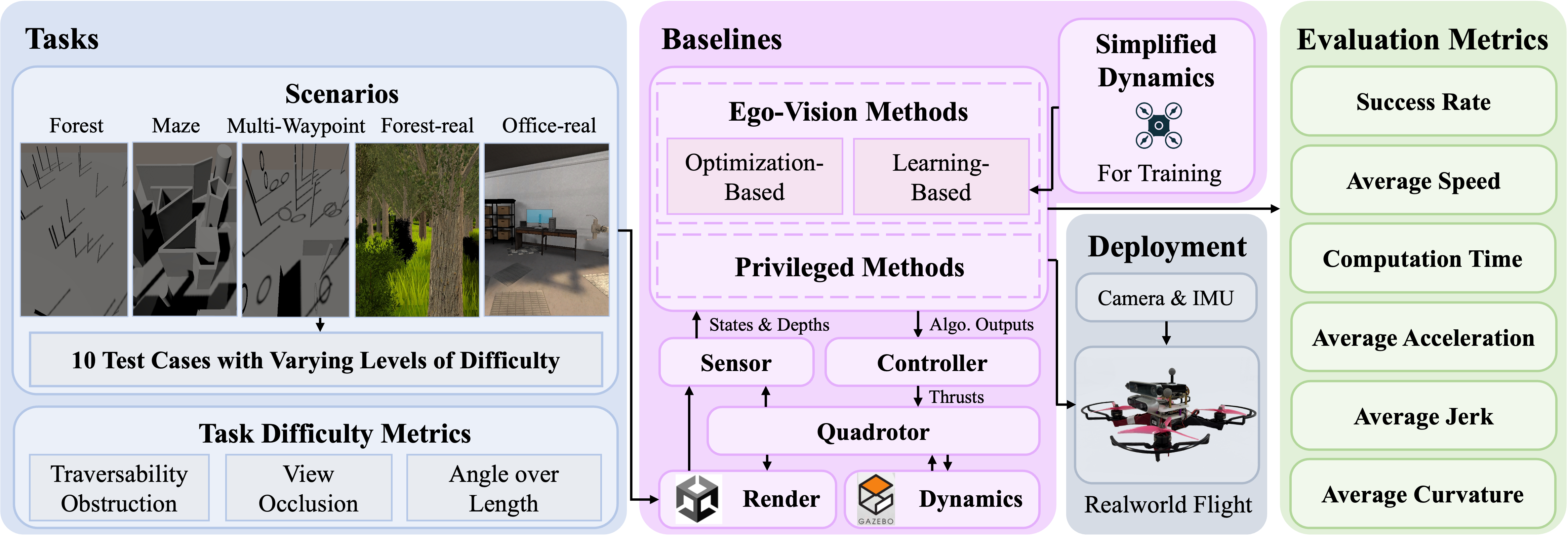}
    \caption{An overview of the FlightBench. FlightBench consists of three main components: (1) Tasks, featuring three scenarios categorized into eight difficulty levels. (2) Baselines, the core benchmarking platform supporting five ego-vision-based {methods} and two privileged {methods}. (3) Evaluation Metrics, offering a thorough suite of performance assessment metrics.}
    \label{fig:overview_flightbench}
    \vspace{-1mm}
\end{figure*}

This section outlines the components of FlightBench, as summarized in \cref{fig:overview_flightbench}. 
To design a set of \texttt{Tasks} with distinguishable characteristics for {assessment}, we propose three criteria, named task difficulty metrics, and develop ten tests across three scenarios based on these metrics. We integrate various representative \texttt{Baselines} to examine the strengths and features of both learning-based and optimization-based navigation methods. Furthermore, we establish a comprehensive set of performance \texttt{Evaluation Metrics} to facilitate quantitative comparisons. The next subsections provide an in-depth look at the Tasks, Baselines, and Evaluation Metrics.

\subsection{Tasks}
\subsubsection{Task Difficulty Metrics}

\begin{figure*}[htbp]
    \centering
    \vspace{1mm}
    \subfigure[Traversability Obstruction]{
        \includegraphics[width = 0.3274\linewidth]{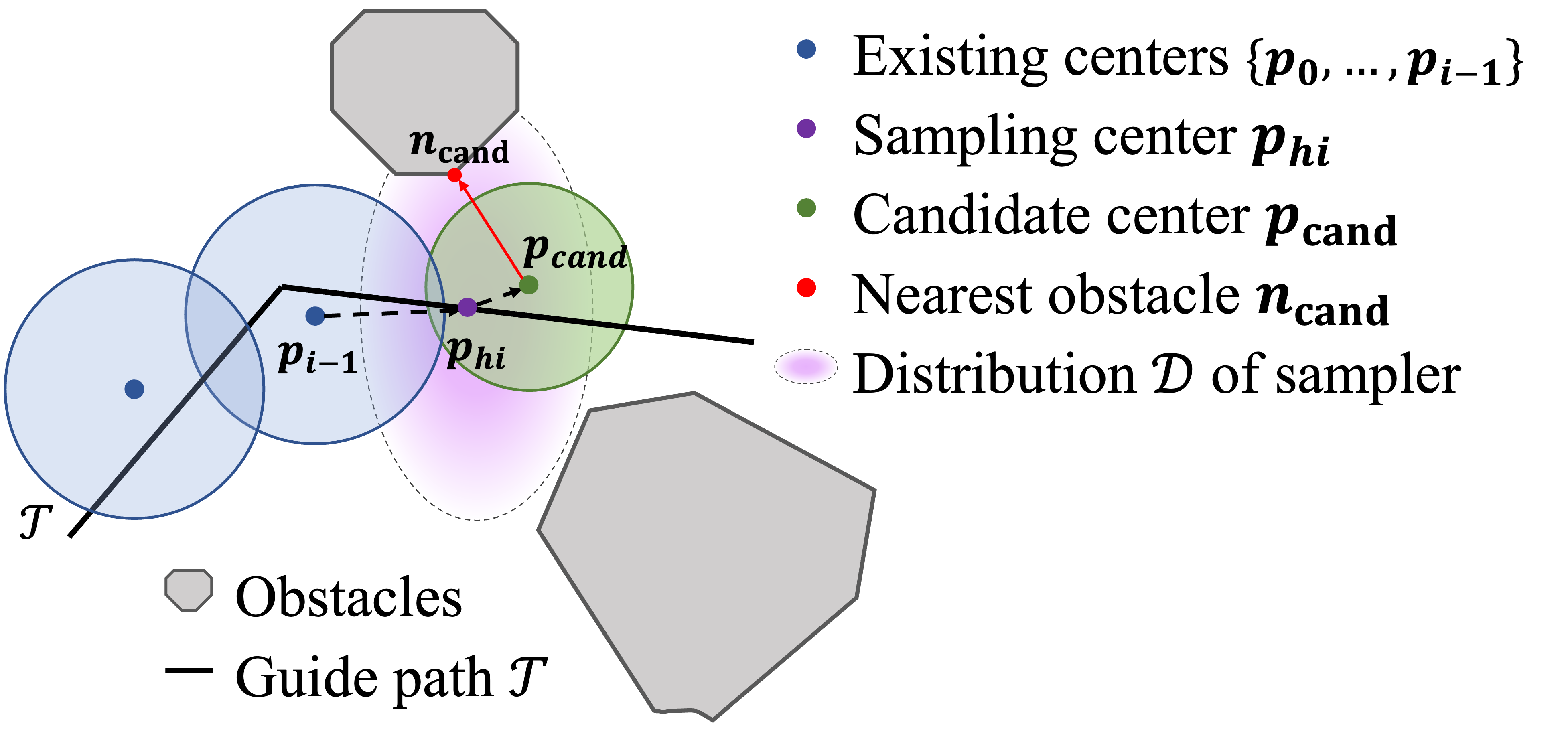}
        \label{fig:TO}
    }
    \subfigure[View Occlusion]{
        \includegraphics[width = 0.3203\linewidth]{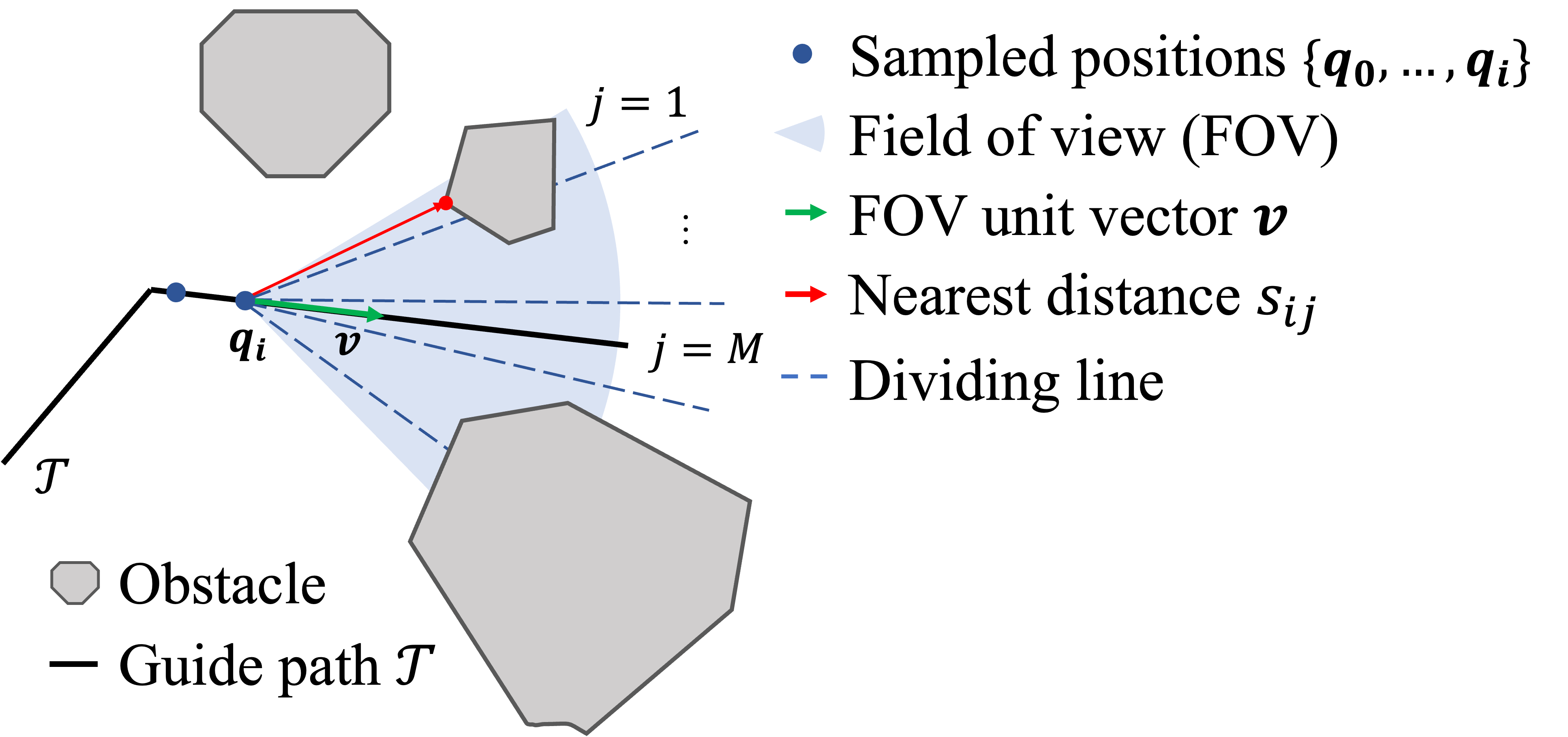}
        \label{fig:VO}
    }
    \subfigure[Angle over Length]{
        \includegraphics[width = 0.2668\linewidth]{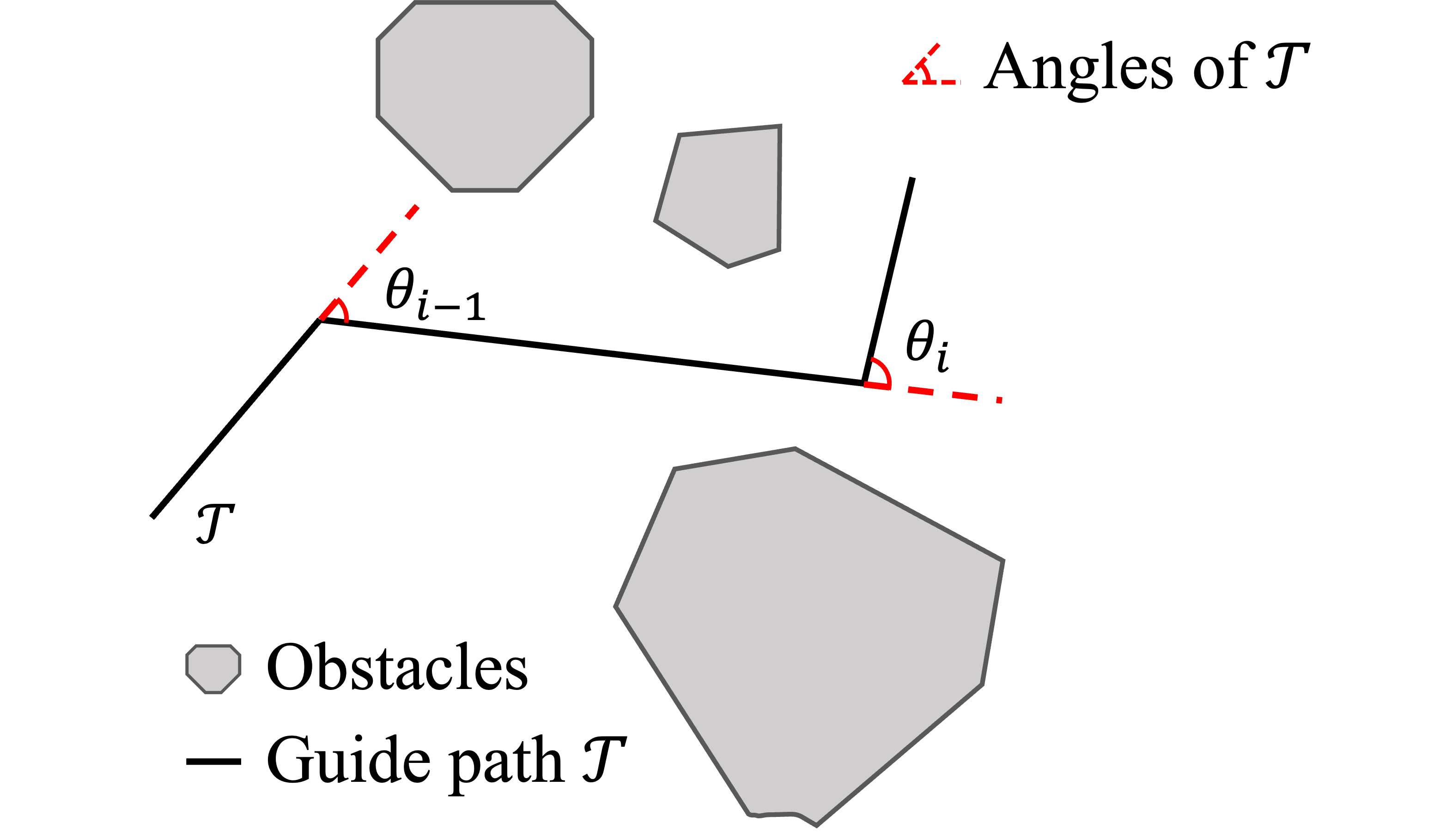}
        \label{fig:AOL}
    } 
    \vspace{-1mm}
    \caption{Illustration of the task difficulty metrics}
    \label{fig:difficulty_metrics}
    \vspace{-2mm}
\end{figure*}
Each task difficulty metric quantifies the challenge of a test configuration from a specific perspective. In quadrotor navigation, a test is configured by the obstacles-laden scenario, start point, and end point.
We utilize the topological guide path $\mathcal{T}$, which comprises interconnected individual waypoints~\cite{penicka2022minimum}, to {establish the quantitative assessment}. In FlightBench, we propose three main task difficulty metrics: {Traversability Obstruction} (TO), {View Occlusion} (VO), and {Angle Over Length} (AOL). 
\paragraph{Traversability Obstruction}
Traversability Obstruction (TO) measures the {flight} difficulty due to limited traversable space caused by obstacles. 
We use a sampling-based approach~\cite{ren2022bubble} to construct sphere-shaped flight corridors $\{B_0, \dots, B_{N_T}\}$, where $N_T$ is the number of spheres representing traversable space along path $\mathcal{T}$. \cref{fig:TO} illustrates the primary notations for computing these corridors.
The next sampling center $\mathbf{p}_\text{hi}$ is chosen from existing spheres $\{B_0, \dots, B_{i-1}\}$ along $\mathcal{T}$. We sample $K$ candidate centers from a 3D Gaussian distribution $\mathcal{D}$ around $\mathbf{p}_\text{hi}$. Each candidate sphere $B_{\text{cand}}$ is defined by its center $\mathbf{p}_\text{cand}$ and radius $r_{\text{cand}} = ||\mathbf{p_\text{cand}} - \mathbf{n_\text{cand}}||_2 - r_d$, where $\mathbf{n_\text{cand}}$ is the nearest obstacle and $r_d$ is the drone radius.
For each $B_{\text{cand}}$, we compute $S_\text{cand}$:
\begin{equation}
S_\text{cand} = k_1 V_\text{cand} + k_2 V_\text{inter} - k_3 (\mathbf{d} \cdot \mathbf{z}) - k_4 ||\mathbf{d} - (\mathbf{d} \cdot \mathbf{z}) \mathbf{z}||_2
\end{equation}
where $k_1, k_2, k_3, k_4 \in \mathbb{R}^+$, $V_\text{cand}$ is the volume of $B_\text{cand}$, $V_\text{inter}$ is the overlap with $B_{i-1}$, $\mathbf{d} = \mathbf{p}_\text{cand} - \mathbf{p}_\text{hi}$, and $\mathbf{z}$ is the unit vector along $\mathbf{p}_\text{hi} - \mathbf{p}_i$. The sphere with the highest $S_\text{cand}$ is selected as the next sphere. This process repeats until path $\mathcal{T}$ is fully covered.
Occlusion challenges mainly occur in narrow spaces, so sphere radii $\{r_1, \dots, r_{N_T}\}$ are sorted in ascending order. The traversability obstruction metric $\mathbb{T}$ is defined in \cref{equ:TO}, where $R$ represents the sensing range.
\begin{equation}
\mathbb{T} = \frac{1}{N_T}\sum_{i=1}^{\lfloor N_T/2 \rfloor} \frac{R}{r_i}.
\label{equ:TO}
\end{equation}

\paragraph{View Occlusion}
In ego-vision-based navigation tasks, a narrow field of view (FOV) can limit the drone's perception, posing a challenge to the perception capabilities of various methods~\cite{tordesillas2022panther, chen2024apace}. We use the term view occlusion (VO) to describe the extent to which obstacles block the FOV in a given scenario. The more obstructed the view, the higher the view occlusion.
As shown in \cref{fig:VO}, we sample drone position $\left\{\mathbf{q_i}\right\}$ and FOV unit vector $\left\{\mathbf{v_i}\right\}$ along $\mathcal{T}$ with $i \in \left\{1,\ \cdots,\ N_V \right\}$. For each sampled pair $\left\{\mathbf{q_i}, \mathbf{v_i}\right\}$, we divide FOV into $M$ parts and calculate the distance $s_{ij}$ between the nearest obstacle point and drone position $q_i$ in each part $j$. The view occlusion $\mathbb V$ can be represented as \cref{equ:vo}, where $m_j$ is a series of weights, which gives higher weight to obstacles closer to the center of the view. 
\begin{equation}
\mathbb{V} = \frac{1}{N_V}\sum_{i=1}^{N_V} \sum_{j=1}^{M} m_j \frac{R}{s_{ij}}.
\label{equ:vo}
\end{equation}
\paragraph{Angle Over Length}
For a given scenario, frequent and violent turns in traversable paths pose challenges for the agility of planning.
We employ the concept of Angle Over Length (AOL) denoted as $\mathbb{A}$ to quantify the sharpness of a path. The AOL $\mathbb{A}$ is defined by \cref{equ:aol}, where $N_{AOL}$ signifies the number of angles depicted in \cref{fig:AOL}, $\theta_i$ represents the $i$-th angle within the topological path $\mathcal{T}$, and $L$ stands for the length of $\mathcal{T}$.
\begin{equation}
    \mathbb{A} = \frac{1}{L}\sum_{i=1}^{N_{AOL}} \left(\exp{\left(\frac{\theta_i}{\pi / 6}\right)}-1\right).
    \label{equ:aol}
\end{equation}

\subsubsection{Scenarios and Tests}
\label{sec:scenarios}
\begin{table}[htbp]
    \caption{Task difficulty score of each test case.}
    \centering
    \begin{tabular}{ccccc}
        \toprule
            Scenarios & Test Cases & TO & VO & AOL\\ \midrule
            \multirow{3}{*}{Forest} & 1 & 0.76 & 0.30 & 7.64$\times$10$^{-4}$ \\
            & 2 & 0.92 & 0.44 & 1.62$\times$10$^{-3}$ \\  
            & 3 & 0.90 & 0.60 &  5.68$\times$10$^{-3}$ \\ \midrule
            \multirow{3}{*}{Maze} & 1 & 1.42 & 0.51 & 1.36$\times$10$^{-3}$ \\
            & 2 & 1.51 & 1.01 & 0.010 \\
            & 3 & 1.54 & 1.39 & 0.61 \\ \midrule
            \multirow{2}{*}{MW}& 1 & 1.81 & 0.55 & 0.08 \\
            & 2 & 1.58 & 1.13 & 0.94 \\ 
            \midrule
            \multirow{1}{*}{Forest-real}& 1 & 1.76 & 1.47 & 0.02 \\
            \midrule
            \multirow{1}{*}{Office-real}& 1 & 0.76 & 0.58 & 8.45$\times 10 ^{-3}$ \\
            \bottomrule
        \end{tabular}
    \label{tab:scenario_metric}
    \vspace{-2mm}
\end{table}
As illustrated in \cref{fig:overview_flightbench}, our benchmark incorporates specific tests based on five scenarios: \texttt{Forest}, \texttt{Maze}, \texttt{Multi-Waypoint}, \texttt{Forest-real}, and \texttt{Office-real}. These scenarios were chosen for their representativeness and frequent use in evaluating quadrotor navigation methods~\cite{ren2022bubble, kaufmann2023champion}. Within these scenarios, we developed ten tests, each characterized by varying levels of difficulty. The task difficulty scores for each test are detailed in \cref{tab:scenario_metric}.
 
The \texttt{Forest} scenario serves as the most common benchmark for quadrotor navigation. We differentiate task difficulty based on obstacle density and establish three tests, following the settings used by Agile\cite{loquercio2021learning}. 
In the \texttt{Forest} scenario, TO and VO metrics increase with higher tree density. AOL is particularly low due to sparsely spanned obstacles, making this scenario suitable for high-speed flights \cite{loquercio2021learning, ren2022bubble}.

The \texttt{Maze} scenario consists of walls and boxes, creating consecutive sharp turns and narrow gaps. Quadrotors must navigate these confined spaces while maintaining flight stability and perception awareness \cite{park2023dlsc}. We devise three tests with varying lengths and turn complexities for \texttt{Maze}, resulting in discriminating difficulty levels for VO and AOL.

The \texttt{Multi-Waypoint} (\texttt{MW}) scenario involves flying through multiple waypoints at different heights sequentially \cite{song2021autonomous}. This scenario also includes boxes and walls as obstacles. The \texttt{MW} scenario is relatively challenging, featuring the highest TO in test 1 and the highest AOL in test 2.

We create two more realistic scenarios: \texttt{Forest-real} and \texttt{Office-real}. The \texttt{Forest-real} scenario features a dense arrangement of textured trees and shrubs on a grassy field. The structure of the obstacles is similar to \texttt{Forest} but with higher density. The \texttt{Office-real} has a structure similar structure to \textit{Maze}, incorporating shelves, desks, and chairs to form narrow gaps and sharp turns. Both scenarios also feature realistic lighting conditions, such as outdoor natural lighting for \texttt{Forest-real} and indoor overhead lighting for \texttt{Office-real}.

\subsection{Baselines}
\begin{figure}
    \centering
    \vspace{2mm}
    \includegraphics[width = 1.0\linewidth]{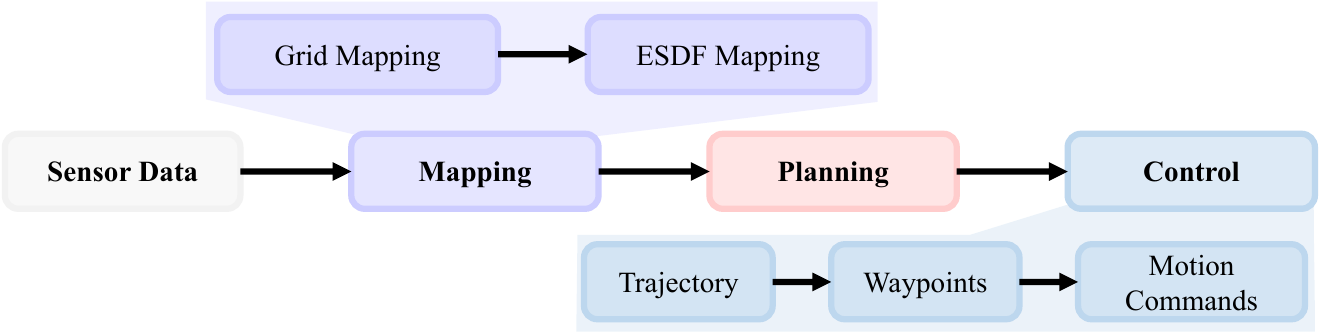}
    \vspace{-2mm}
    \caption{A generic processing pipeline for hierarchical navigation systems on quadrotors. }
    \vspace{-4mm}
    \label{fig:baseline}
\end{figure}

We evaluate representative vision-based navigation methods in FlightBench, including three learning-based approaches (\textbf{Agile}~\cite{loquercio2021learning}, \textbf{NPE}~\cite{zhang2024npe}, \textbf{LPA}~\cite{song2023learning}) and three optimization-based approaches (\textbf{Fast-Planner}~\cite{zhou2019robust}, \textbf{TGK-Planner}~\cite{ye2020tgk}, \textbf{EGO-Planner}~\cite{zhou2020ego}). To establish an upper bound, we also integrate two privileged methods with access to environmental information (\textbf{SBMT}~\cite{penicka2022minimum}, \textbf{LMT}~\cite{penicka2022learning}). We offer two perception module options: ground-truth depth and SGM depth~\cite{hirschmuller2007stereo}. These methods produce varying outputs, which are then processed by the same control stack (see \cref{fig:baseline}) with an MPC controller~\cite{falanga2018pampc}. The characteristics of each method are detailed in \cref{tab:planner_properties}. For code, parameters, and implementation details, please visit our website\footnote{\label{footnote:website} \url{https://thu-uav.github.io/FlightBench}}.

\begin{table*}[htbp]
    \centering
    \vspace{1mm}
    \caption{{Characteristics of the {navigation methods for quadrotors}. "RL" denotes reinforcement learning. "IL" represents imitation learning.
    "GM" and "EM" refer to Grid Mapping and ESDF Mapping, respectively. The control level indicates the part of the control stack used by the baseline.}}
    \begin{tabular}{c|ccc|ccccc}
    \toprule
        \multirow{2}{*}{~} & \multirow{2}{*}{Method Type} & Priv. & Decision & \multirow{2}{*}{Mapping} & \multirow{2}{*}{Planning}  & \multicolumn{3}{c}{Control Stack} \\ 
        & &Info. &Horizon & & &Traj. &Waypoint &Motion Cmd \\ \midrule
        SBMT & Samp.-based &\cmark &Global &\multicolumn{4}{c}{\cellcolor{customblue!50}Planning Module} &\multicolumn{1}{c}{\cellcolor{customyellow!50}MPC} \\ 
        LMT & RL &\cmark & Local &\multicolumn{5}{c}{\cellcolor{customblue!50}Policy Network} \\ 
        \cmidrule{1-8}
        Fast-Planner &Opti.-based &\xmark &Global &\cellcolor{customgreen!50}{GM+EM} &\multicolumn{2}{c}{\cellcolor{customblue!50}Planning Module} & \multicolumn{2}{c}{\cellcolor{customyellow!50}MPC} \\ 
        EGO-Planner &Opti.-based &\xmark &Local &\cellcolor{customgreen!50}{GM} &\multicolumn{2}{c}{\cellcolor{customblue!50}Planning Module} & \multicolumn{2}{c}{\cellcolor{customyellow!50}MPC} \\ 
        TGK-Planner &Opti.-based &\xmark &Global &\cellcolor{customgreen!50}{GM} &\multicolumn{2}{c}{\cellcolor{customblue!50}Planning Module} & \multicolumn{2}{c}{\cellcolor{customyellow!50}MPC} \\ 
         Agile & IL &\xmark &Local &\multicolumn{4}{c}{\cellcolor{customblue!50}Policy Network} & \multicolumn{1}{c}{\cellcolor{customyellow!50}MPC} \\
        LPA & RL+IL &\xmark &Local & \multicolumn{5}{c}{\cellcolor{customblue!50}Policy Network} \\ 
        NPE & RL & \xmark & Local & \multicolumn{4}{c}{\cellcolor{customblue!50}Policy Network} & \multicolumn{1}{c}{\cellcolor{customyellow!50}MPC}\\
        \bottomrule
    \end{tabular}
    \label{tab:planner_properties}
    \vspace{-2mm}
\end{table*}

\subsection{Evaluation Metrics}
\label{sec:eval_metrics}
We represent quadrotor states as a tuple $\left(\mathbf{x}(t),\mathbf{v}(t),\mathbf{a}(t), \mathbf{j}(t)\right)$,  where $t$ denotes time, $\mathbf{x}(t)$ denotes the position, and $\mathbf{v}(t) = \dot{\mathbf{x}}(t)$, $\mathbf{a}(t) = \dot{\mathbf{v}}(t)$, $\mathbf{j}(t) = \dot{\mathbf{a}}(t)$ are the velocity, acceleration, and jerk in the world frame, respectively. $T$ denotes the time taken to fly from the starting point to the end point. 

First, we integrate three widely used metrics~\cite{ren2022bubble, loquercio2021learning} into FlightBench. \textbf{Success rate} measures if the quadrotor reaches the goal within a 1.5 m radius without crashing. \textbf{Average speed}, defined as $\frac{1}{T} \int_0^T ||\mathbf{v}(t)||_2 \, \mathrm{d}t$, reflects the achieved agility. \textbf{Computation time} evaluates real-time performance as the sum of processing times for mapping, planning, and control. Additionally, we introduce \textbf{average acceleration} and \textbf{average jerk} \cite{zhou2019robust, zhou2020ego}, defined as $\bar{\mathbf{a}} = \frac{1}{L} \int_0^T ||\mathbf{a}(t)||_2^2 \, \mathrm{d}t$ and $\bar{\mathbf{j}} = \frac{1}{L} \int_0^T ||\mathbf{j}(t)||_2^2 \, \mathrm{d}t$, respectively. Average acceleration indicates energy consumption, while average jerk measures flight smoothness \cite{wang2022geometrically}. These metrics, though not commonly used in evaluations, are crucial for assessing practicability and safety in real-world applications.
However, average acceleration and jerk only capture the dynamic characteristics of a flight. For instance, higher flight speeds along the same trajectory result in greater average acceleration and jerk. To assess the static quality of trajectories, we propose \textbf{average curvature}, inspired by Bench-MR~\cite{heiden2021bench}. Curvature is calculated as $\kappa(t) = \frac{|\mathbf{v}(t) \times \mathbf{a}(t)|}{|\mathbf{v}(t)|^3}$, and average curvature is defined as $\bar{\kappa} = \frac{1}{L} \int_0^T \kappa(t) \mathbf{v}(t) \, \mathrm{d}t$.

Together, these six metrics provide a comprehensive comparison of {learning-based algorithms against optimization-based methods for ego-vision-based quadrotor navigation}. Our extensive experiments will demonstrate that while learning-based methods excel in certain metrics, they also have shortcomings in others.

%% file: 5_exp.tex
\section{Experiments}

Using FlightBench, we carried out extensive experiments to address the following research questions:
\begin{itemize}
    \item What are the main advantages and limitations of learning-based methods compared to optimization-based methods?
    \item How does navigation performance vary across different scenario settings?
    \item How much does the system latency introduced by the practical evaluation environment affect performance?
    \item Is there a consistent correlation between performance in simulation and those demonstrated in real-world?
\end{itemize}

\subsection{Setup}
\label{sec:settings}
For simulating quadrotors, we use Flightmare~\cite{song2020flightmare} with Gazebo~\cite{gazebo} as its dynamic engine. To mimic real-world conditions, we develop a simulated quadrotor model equipped with an IMU sensor and a depth camera, calibrated with real flight data. To simulate real-world communication delays, all data transmission in the simulation uses ROS~\cite{quigley2009ros}.
All simulations are conducted on a desktop PC with an Intel Core i9-11900K processor and an Nvidia 3090 GPU. Each evaluation metric is averaged over ten independent runs. 

We validate the sim-to-real capabilities of our benchmark through real-world experiments on the Q250 platform. Q250 quadrotor integrates an NVIDIA Orin NX for computation and utilizes MAVLink to interface with the PX4 flight controller for low-level control.
The implementation details and full experimental results are available on our website\ref{footnote:website}.

\subsection{Benchmarking Flight Performance}
\label{sec:performance}
\begin{table*}[t]
    \vspace{2mm}
    \centering
    \caption{Performance evaluation of different methods under tests with the highest AOL within each scenario. The highest performing values for each metric are highlighted in bold, with the second highest underlined.}
    \begin{tabular}{cr|cc|ccc|ccc}
        \toprule
        \multirow{2}{*}{Scen.} &  \multicolumn{1}{c}{\multirow{2}{*}{Metric}} &\multicolumn{2}{c|}{Privileged} &\multicolumn{3}{c|}{Optimization-based} &\multicolumn{3}{c}{Learning-based} \\
        & & SBMT & LMT & TGK & Fast & EGO & Agile & LPA & NPE\\ \midrule
        \multirow{5}{*}{Forest} & Success Rate $\uparrow$ & 0.80 & \textbf{1.00} & \underline{0.90} & \underline{0.90} & \textbf{1.00} & \underline{0.90} & \textbf{1.00} & \underline{0.90} \\
        ~ & Avg. Spd. (ms$^{-1}$) $\uparrow$ & \textbf{15.25} & \underline{11.84} & 2.30	& 2.47 & 2.49  & 3.058 & 8.96 & 2.03 \\
        ~ & Avg. Curv. (m$^{-1}$) $\downarrow$ & \textbf{0.06}  & \underline{0.07}  & 0.08  & \textbf{0.06}  & 0.08  & 0.37  & 0.08 & 0.13 \\
        ~ & Avg. Acc. (ms$^{-3}$) $\downarrow$ & 28.39 & 10.29 & \underline{0.25} & \textbf{0.19} & 0.83 & 4.93 & 9.96 & 0.98\\
        ~ & Avg. Jerk (ms$^{-5}$) $\downarrow$& 4.27$\times$10$^3$ & 8.14$\times$10$^3$ & \textbf{1.03} & \underline{3.97} & 58.39 & 937.02 & 1.14$\times$10$^4$ & 1.82 $\times 10^3$ \\ \midrule
        \multirow{5}{*}{Maze} & Success Rate $\uparrow$ & \underline{0.60} & \textbf{0.9} & 0.50 & \underline{0.60} & 0.20 & 0.50 & 0.30 & 0.50 \\
        ~ & Avg. Spd. (ms$^{-1}$) $\uparrow$ & \underline{8.73} & \textbf{9.62}  & 1.85 & 1.99 & 2.19 & 3.00 & 8.35 & 1.98\\
        ~ & Avg. Curv. (m$^{-1}$) $\downarrow$ & 0.31 & \textbf{0.13} & \underline{0.17} & 0.23 & 0.33 & 0.68 & 0.21 & 0.39 \\
        ~ & Avg. Acc. (ms$^{-3}$) $\downarrow$ & 60.73 & 26.26 & \textbf{0.50} & \underline{0.79} & 1.91 & 15.45 & 37.30 & 2.12\\
        ~ & Avg. Jerk (ms$^{-5}$) $\downarrow$ & 6.60$\times$10$^3$ & 4.64$\times$10$^3$ & \textbf{6.74} & \underline{9.62} & 80.54 & 2.15$\times$10$^3$ & 4.64$\times$10$^3$ & 2.38$\times 10^3$\\ \midrule
        \multirow{5}{*}{MW} & Success Rate $\uparrow$ & 0.70 & \textbf{0.90} & 0.40 & \underline{0.80} & 0.50 & 0.60 & 0.50 & 0.40\\
        ~ & Avg. Spd. (ms$^{-1}$) $\uparrow$ & 5.59 & \textbf{6.88} & 1.48 & 1.73 & 2.13 & 3.05 & \underline{6.72} & 1.50\\
        ~ & Avg. Curv. (m$^{-1}$) $\downarrow$ & 0.47  & \underline{0.30}  & 0.46  & 0.32  & 0.62  & 0.67  & \textbf{0.26} & 0.71 \\
        ~ & Avg. Acc. (ms$^{-3}$) $\downarrow$ & 80.95 & 31.23 & \underline{1.07} & \textbf{0.97} & 5.06 & 16.86 & 36.77 & 2.84 \\
        ~ & Avg. Jerk (ms$^{-5}$) $\downarrow$ & 9.76$\times$10$^3$ & 1.66$\times$10$^4$ & \underline{25.52} & \textbf{22.72} & 155.83 & 2.07$\times$10$^3$ & 6.19$\times$10$^3$ & 2.53$\times 10^3$ \\
        \midrule
        \multirow{5}{*}{Forest-real} & Success Rate $\uparrow$ & \underline{0.90} & \textbf{1.00} & 0.70 & \underline{0.90} & \underline{0.90} & \textbf{1.00} & \underline{0.90} & \underline{0.90} \\
        ~ & Avg. Spd. (ms$^{-1}$) $\uparrow$ & \textbf{14.11} & \underline{11.20} & 2.40 & 2.42 & 2.44 & 3.06 & 9.70 & 2.16 \\
        ~ & Avg. Curv. (m$^{-1}$) $\downarrow$ & \textbf{0.04}  & 0.10  & \underline{0.09}  & \underline{0.09}  & 0.26  & 0.24  & 0.13 & 0.15 \\
        ~ &Avg. Acc. (ms$^{-3}$) $\downarrow$ & 28.56 & 10.90 & \textbf{0.18} & \underline{0.29} & 1.82 & 4.87 & 16.49 & 1.92 \\
        ~ & Avg. Jerk (ms$^{-5}$) $\downarrow$ & 4.37$\times$10$^3$ & 8.20$\times$10$^3$ & \underline{3.09} & \textbf{2.25} & 94.70 & 862.3 & 6.72$\times$10$^3$ & 1.89$\times 10^3$ \\
        \midrule
        \multirow{5}{*}{Office-real} & Success Rate $\uparrow$ & \underline{0.80} & \textbf{0.90} & 0.70 & 0.70 & 0.50 & 0.50 & \underline{0.80} & 0.60 \\
        ~ & Avg. Spd. (ms$^{-1}$) $\uparrow$ & \underline{9.54} & \textbf{10.45} & 1.86 & 2.01 & 2.51 & 3.01 & 9.32 & 1.95 \\
        ~ & Avg. Curv. (m$^{-1}$) $\downarrow$ & \textbf{0.06}  & \underline{0.09}  & 0.19  & 0.18  & 0.25  & 0.33  & \underline{0.09} & 0.36 \\
        ~ & Avg. Acc. (ms$^{-3}$) $\downarrow$ & 39.08 & 11.24 & \textbf{0.49} & \underline{0.59} & 3.10 & 8.41 & 20.64 & 3.38 \\
        ~ & Avg. Jerk (ms$^{-5}$) $\downarrow$ & 5.40$\times$10$^3$ & 4.52$\times$10$^3$ & \textbf{7.28} & \underline{9.16} & 285.91 & 1.03$\times$10$^3$ & 7.11$\times$10$^3$ & 1.82$\times 10^3$ \\
        \bottomrule
    \end{tabular}
    \label{tab:quality}
    \vspace{-1mm}
\end{table*}
\subsubsection{Flight Quality}
To systematically assess the strengths and weaknesses of various methods on ego-vision navigation, we conduct evaluations across all tests within five distinct scenarios. \cref{tab:quality} displays the results for tests with the highest AOL in each scenario. We evaluate these methods using our proposed evaluation metrics, and computation time will be addressed separately in a subsequent discussion. 
In these experiments, we standardize the expected maximum speed at $3m/s$ for a fair comparison. Exceptions are SBMT, LMT, and LPA, whose flight speeds cannot be manually controlled.

As shown in \cref{tab:quality}, the privileged methods, with a global awareness of obstacles, set the upper bound for motion performance in terms of average speed and success rate. In contrast, the success rate of ego-vision methods in the Maze and MW scenarios is generally below $0.6$, indicating that our benchmark remains challenging for ego-vision methods, especially at the perception level.

Learning-based methods, known for their aggressive maneuvering, tend to fly less smoothly and consume more energy. They also experience more crashes in areas with large corners, as seen in the Maze and MW scenarios. When performing a large-angle turn, an aggressive policy is more likely to cause the quadrotor to lose balance and crash. Optimization-based methods are still competitive or even superior to current learning-based approaches, particularly in terms of minimizing energy costs. By contrasting the more effective Fast-Planner with the more severely impaired TGK-Planner and EGO-Planner, we find that global trajectory smoothing and enhancing the speed of replanning are crucial for improving success rates in complex scenarios.

\begin{remark}
    Learning-based methods tend to execute aggressive and fluctuating maneuvers, yet they struggle with instability in scenarios with high challenges related to VO and AOL. 
\end{remark}

\subsubsection{Impact of Flight Speed and Computation Time}
\label{sec:flightspeed}

\begin{figure}[htbp]
    \centering
    \includegraphics[width=0.8\linewidth]{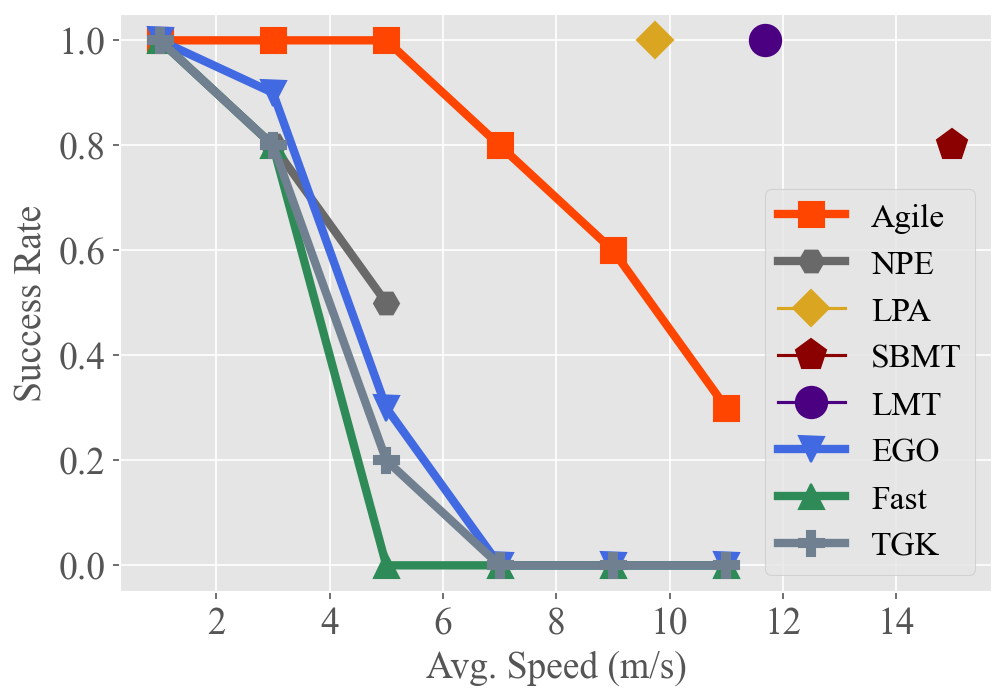}
    \vspace{-1mm}
    \caption{Success rate with different flight speeds.}
    \label{fig:max_speed}
    \vspace{-3mm}
\end{figure}

This section evaluates all methods in Test Case 2 of the Forest scenario. As illustrated in \cref{fig:max_speed}, we examine the success rate of each method at different average speeds, excluding three methods where speed is non-adjustable. Learning-based methods exhibit agile obstacle avoidance and operate near dynamic limits due to their end-to-end architecture. In contrast, optimization-based methods struggle at high speeds, as their hierarchical pipeline latency often causes the quadrotor to overshoot obstacles before a new path is generated. Privileged methods achieve higher success rates and faster flights, highlighting the potential for improving current ego-vision-based approaches.

We also analyze computation times for various planning methods on both desktop and onboard platforms, breaking down the results into mapping, planning, and control stages (see \cref{tab:inference_time}, \cref{fig:baseline}). For learning-based methods, mapping time includes image-to-tensor conversion and state pre-processing. Because of the compact neural network structure, learning-based methods generally require less computation time. Among optimization-based approaches, mapping is the most time-intensive stage, particularly on the onboard platform. Fast-Planner has the longest mapping time due to ESDF map construction. Similar to the conclusions of \cite{loquercio2021learning}, computation time directly impacts maximum flight speed. Lighter architectures, which allow higher replan frequencies, improve obstacle reaction and enable faster flights.

\begin{table*}[htbp]
    \vspace{2mm}
    \caption{Computation time of different baselines. $T_{\text{map}},\ T_{\text{plan}},\ T_{\text{ctrl}},\ T_{\text{tot}}$  stands for mapping time, planning time, control time, and total time, respectively.}
    \centering
    \begin{tabular}{c|c|cccccccc}
    \toprule
        \multicolumn{2}{c}{} & SBMT & LMT & TGK & Fast & EGO & Agile & LPA & NPE\\ \midrule
        \multirow{4}{*}{Desktop} & \cellcolor{gray!20}$T_{\text{tot}}$ (ms) & \cellcolor{gray!20}3.189$\times$10$^{5}$ & \cellcolor{gray!20}2.773 & \cellcolor{gray!20}11.960  & \cellcolor{gray!20}8.196 & \cellcolor{gray!20}3.470  & \cellcolor{gray!20}5.573 & \cellcolor{gray!20}1.395 & \cellcolor{gray!20}4.942\\
        ~ & $T_{\text{map}}$ (ms) & -  & 1.607  & 3.964  & 7.038  & 2.956  & 0.338  & 0.399 & 2.107  \\ 
        ~ & $T_{\text{plan}}$ (ms) & 2.589$\times$10$^{5}$ & 1.167   & 7.994 & 1.155 & 0.510  & 5.115  & 0.995  & 2.833\\ 
        ~ & $T_{\text{ctrl}}$ (ms) & - & -  & 0.002 & 0.003 & 0.003 & 0.119  & - & 0.003\\
        \midrule
        \multirow{4}{*}{Onboard} & \cellcolor{gray!20}$T_{\text{tot}}$ (ms) & \cellcolor{gray!20}- & \cellcolor{gray!20}13.213 & \cellcolor{gray!20}37.646 & \cellcolor{gray!20}39.177 & \cellcolor{gray!20}24.946 & \cellcolor{gray!20}27.458 & \cellcolor{gray!20}12.293 &
        \cellcolor{gray!20}17.144\\
        ~ & $T_{\text{map}}$ (ms) & - &  2.768 & 27.420 & 36.853 & 24.020 & 1.175 & 4.313 & 4.532 \\
        ~ & $T_{\text{plan}}$ (ms)  & - & 10.445 & 10.211 & 2.310 & 0.910 & 26.283 & 7.980 & 12.598 \\ 
        ~ & $T_{\text{ctrl}}$ (ms) & - &  - & 0.015 & 0.014 & 0.016 & - & - & 0.014\\ 
        \bottomrule
    \end{tabular}
    \label{tab:inference_time}
    \vspace{-3mm}
\end{table*}

\begin{remark}
    Learning-based methods, with their compact neural network architectures, excel in high-speed flight by achieving superior performance and faster planning. Well-crafted optimization-based methods can achieve comparable planning speed.
\end{remark}

\begin{table*}[htbp]
    \centering
    \vspace{2mm}
    \caption{HITL experiment results \label{tab:hitl}}
    \scalebox{0.92}{
    \begin{tabular}{c|ccc|cccc|cccc}
    \toprule
    \multirow{3}{*}{Scene} & \multirow{3}{*}{TO} & \multirow{3}{*}{VO} & \multirow{3}{*}{AOL} & \multicolumn{4}{c}{EGO} & \multicolumn{4}{c}{Agile} \\
    & ~ & ~ & ~ & \makecell[c]{Avg. Spd. $\uparrow$ \\(ms$^{-1}$)} & \makecell[c]{Avg. Curv. $\downarrow$ \\(m$^{-1}$)} & \makecell[c]{Avg. Acc. $\downarrow$ \\(ms$^{-3}$)} & \makecell[c]{Avg. Jerk $\downarrow$ \\(ms$^{-5}$)} & \makecell[c]{Avg. Spd. $\uparrow$ \\(ms$^{-1}$)} & \makecell[c]{Avg. Curv. $\downarrow$ \\(m$^{-1}$)} & \makecell[c]{Avg. Acc. $\downarrow$ \\(ms$^{-3}$)} & \makecell[c]{Avg. Jerk $\downarrow$ \\(ms$^{-5}$)} \\
    \midrule
    1 & 0.98 & 0.57 & 3$\times 10^{-4}$ & 1.39 & 0.54 & 1.40 & 4.85$\times 10^2$ & 1.45 & 0.65 & 2.11 & 3.99$\times 10^3$\\
    2 & 1.81 & 1.82 & 0.027 & 1.32 & 0.92 & 1.68 & 5.64$\times 10^2$ & 1.47 & 1.61 & 5.62 & 6.64$\times 10^3$ \\
    3 & 2.07 & 1.85 & 0.040 & 1.39 & 1.55 & 5.20 & 7.70$\times 10^2$ & 1.51 & 1.62 & 7.82 & 8.36 $\times 10^3$\\
    \bottomrule
    \end{tabular}}
    \vspace{-2mm}
\end{table*}

\subsection{Analyses on Effectiveness of Different Metrics}
\label{sec:corr}
\begin{figure}[htbp]
    \centering
    \vspace{-1mm}
    \subfigure[Correlation heatmap for privileged methods]{
        \includegraphics[width=0.75\linewidth]{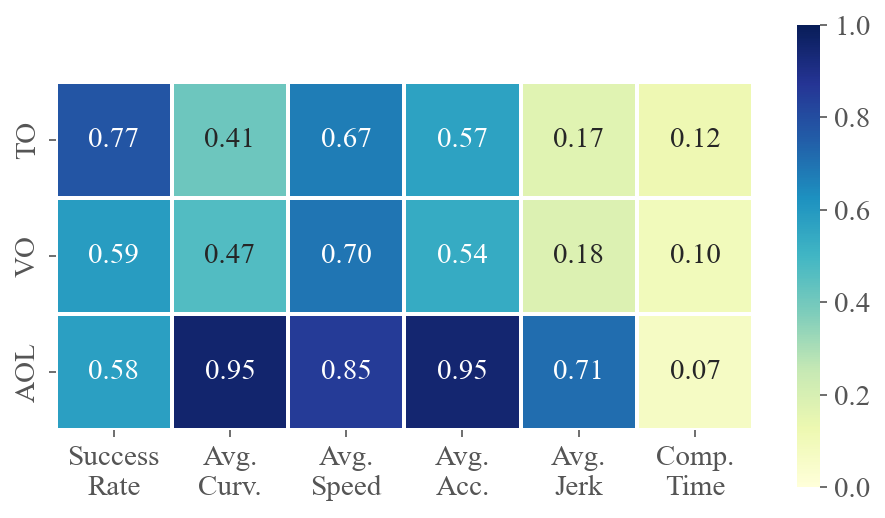}
        \label{fig:corr_priv}
    }
    \subfigure[Correlation heatmap for ego-vision methods]{
        \includegraphics[width=0.75\linewidth]{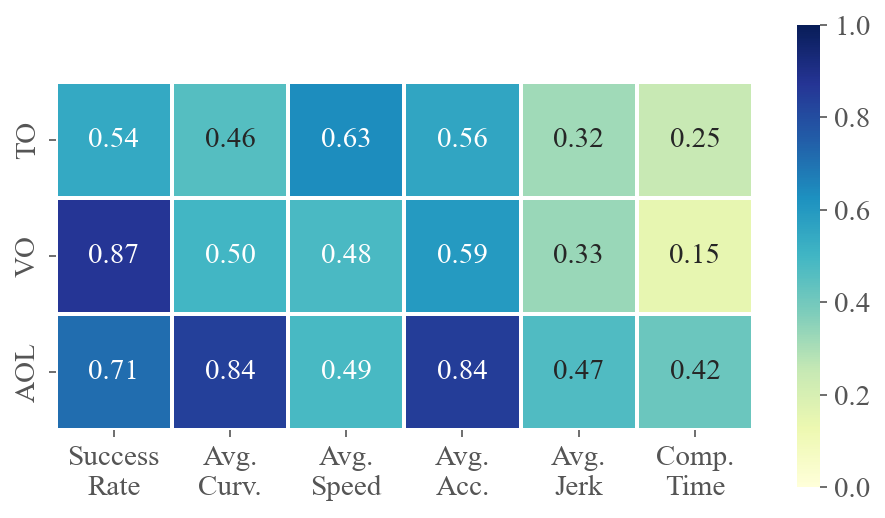}
        \label{fig:corr_vis}
    }
    \caption{Correlation coefficients between three difficulty metrics and six evaluation metrics.}
    \label{fig:corr}
\end{figure}

To demonstrate how various task difficulties influence different aspects of flight performance, we calculate the correlation coefficients between six performance metrics and three difficulty metrics for each method across multiple {tests}. The value at the intersection of the horizontal and vertical axes represents the absolute value of the correlation coefficient between the two metrics. A higher value indicates a stronger correlation.
\cref{fig:corr} presents the average correlations for privileged and ego-vision-based methods, separately evaluating the impacts on agility and partial observation.

The results for privileged methods shown in \cref{fig:corr_priv} indicate that AOL and TO have a significant impact on the baseline's motion performance. The correlation coefficients between AOL and average curvature, velocity and acceleration are all above 0.85, indicating that AOL describes the sharpness of the trajectory well. More specifically, high AOL results in high curvature and acceleration of the flight trajectory, as well as lower average speed. TO, indicating task narrowness, is a crucial determinant of flight success rates. In contrast to privileged methods where global information is accessible, as shown in \cref{fig:corr_vis}, ego-vision-based methods primarily struggle with partial perception. Field-of-view occlusions and turns challenge real-time environmental awareness, making VO highly correlated with success rates.
\begin{remark}
    High VO and AOL significantly challenge learning-based methods, as these factors heavily impact the ego-vision-based method's ability to handle partial observations and sudden reactions.
\end{remark}

\subsection{Impact of Latency on Learning-based Methods}
\label{sec:latency}
\begin{table}[htbp]
    \centering
    \caption{Performance of RL-based methods under test 2 of Multi-Waypoint scenario.}
    \scalebox{0.95}{
    \begin{tabular}{ccccc}
        \toprule
         ~ & \multicolumn{2}{c}{Train w/ latency} & \multicolumn{2}{c}{Train w/o latency} \\ 
         \cmidrule(r){2-5}
         ~ & Success Rate $\uparrow$ & Progress & Success Rate $\uparrow$ & Progress\\ \midrule
        LMT & \textbf{0.9} & \textbf{0.96} & 0.3 & 0.57\\
        LPA & \textbf{0.5} & \textbf{0.73} & 0.0 & 0.32 \\ \bottomrule
    \end{tabular}}
    \label{tab:latency}
    \vspace{-3mm}
\end{table}
Beyond the algorithmic factors previously discussed, learning-based methods face considerable challenges transitioning from training simulations to real-world applications. Latency significantly impacts sim-to-real transfer, particularly when simplified robot dynamics are used to enhance high-throughput RL training. We evaluate the impact of latency in a ROS-based environment, where asynchronous node communication introduces approximately $45ms$ of delay.

\cref{tab:latency} details the performance of RL-based methods under the most challenging test, i.e., test 2 of the Multi-Waypoint scenario, with the highest VO, AOL, and the second-highest TO. To further evaluate the baseline's performance at low success rates, we introduce \emph{Progress}, a metric ranging from 0 to 1 that reflects the proportion of the trajectory completed before a collision occurs. The ``train w/ latency'' column displays results from training with simulated and randomized latencies between $25ms$ to $50ms$, whereas the ``train w/o latency'' column serves as a control group. 
``Train w/ latency'' significantly improves both success rate and progress by more than $50\%$ for two methods, emphasizing the importance of incorporating randomized latency in more realistic simulation environments and real-world deployments.

\begin{remark}
    Integrating latency randomization into the training of RL-based methods is essential to enhance real-world applicability.
\end{remark}

\subsection{Real-world Flight}
\label{sec:realworld}
We selected two representative methods for real-world experiments: EGO, from the optimization-based category, and Agile, from the learning-based category. First, we conducted full-pipeline experiments in cluttered environments to evaluate the deployment capabilities of our benchmark (\cref{fig:realworld}). During these flights, all state estimation, sensing, computation, and control were executed entirely onboard.

Next, we performed Hardware-In-The-Loop (HITL) experiments, where simulated visual perception guided the drone’s flight, allowing us to quantitatively validate the consistency between simulation and real-world performance. \cref{tab:hitl} presents flight results across three real-world test cases. In more challenging scenarios (characterized by higher TO, VO, and AOL), the quadrotor exhibited increased curvature, acceleration, and jerk. Furthermore, in these scenarios, Agile demonstrated greater curvature, acceleration, and jerk compared to EGO, leading to less smooth flight. These findings confirm that the performance trends observed in simulation closely align with real-world outcomes. For further details, please refer to our video.

\begin{remark}
    Our benchmark supports full-pipeline deployment, ensuring that flight performance trends remain consistent and order-preserving across simulation and real-world conditions.
\end{remark}

%% file: 6_con.tex
\section{Conclusion}
\label{sec:conclusion}
In this paper, we introduce FlightBench, an open-source benchmark for evaluating both learning-based and optimization-based approaches in ego-vision quadrotor navigation. By incorporating three task difficulty metrics, FlightBench enables quantitative comparisons across diverse scenarios and test cases. Our experiments demonstrate that learning-based methods excel in high-speed flight and inference efficiency but require improvement in handling complexity and robustness. In contrast, optimization-based methods, while slower, consistently generate smooth and flyable trajectories. Correlation analysis confirms that our task difficulty metrics effectively capture performance variations across different test cases. Additionally, real-world experiments validate the consistency between simulation and actual quadrotor performance. We hope FlightBench will serve as a catalyst for advancing learning-based quadrotor navigation.

%% file: main.bbl
\begin{thebibliography}{10}
\providecommand{\url}[1]{#1}
\csname url@samestyle\endcsname
\providecommand{\newblock}{\relax}
\providecommand{\bibinfo}[2]{#2}
\providecommand{\BIBentrySTDinterwordspacing}{\spaceskip=0pt\relax}
\providecommand{\BIBentryALTinterwordstretchfactor}{4}
\providecommand{\BIBentryALTinterwordspacing}{\spaceskip=\fontdimen2\font plus
\BIBentryALTinterwordstretchfactor\fontdimen3\font minus \fontdimen4\font\relax}
\providecommand{\BIBforeignlanguage}[2]{{%
\expandafter\ifx\csname l@#1\endcsname\relax
\typeout{** WARNING: IEEEtran.bst: No hyphenation pattern has been}%
\typeout{** loaded for the language `#1'. Using the pattern for}%
\typeout{** the default language instead.}%
\else
\language=\csname l@#1\endcsname
\fi
#2}}
\providecommand{\BIBdecl}{\relax}
\BIBdecl

\bibitem{xiao2023collaborative}
J.~Xiao, P.~Pisutsin, and M.~Feroskhan, ``Collaborative target search with a visual drone swarm: An adaptive curriculum embedded multistage reinforcement learning approach,'' \emph{IEEE Transactions on Neural Networks and Learning Systems}, 2023.

\bibitem{xiao2021learning}
X.~Xiao, J.~Biswas, and P.~Stone, ``Learning inverse kinodynamics for accurate high-speed off-road navigation on unstructured terrain,'' \emph{IEEE Robotics and Automation Letters}, vol.~6, no.~3, pp. 6054--6060, 2021.

\bibitem{stachowicz2023fastrlap}
K.~Stachowicz, D.~Shah, A.~Bhorkar, I.~Kostrikov, and S.~Levine, ``Fastrlap: A system for learning high-speed driving via deep rl and autonomous practicing,'' in \emph{Conference on Robot Learning}.\hskip 1em plus 0.5em minus 0.4em\relax PMLR, 2023, pp. 3100--3111.

\bibitem{agarwal2023legged}
A.~Agarwal, A.~Kumar, J.~Malik, and D.~Pathak, ``Legged locomotion in challenging terrains using egocentric vision,'' in \emph{Conference on robot learning}.\hskip 1em plus 0.5em minus 0.4em\relax PMLR, 2023, pp. 403--415.

\bibitem{loquercio2021learning}
A.~Loquercio, E.~Kaufmann, R.~Ranftl, M.~M{\"u}ller, V.~Koltun, and D.~Scaramuzza, ``Learning high-speed flight in the wild,'' \emph{Science Robotics}, vol.~6, no.~59, p. eabg5810, 2021.

\bibitem{xiao2024visionbased}
J.~Xiao, R.~Zhang, Y.~Zhang, and M.~Feroskhan, ``Vision-based learning for drones: A survey,'' 2024.

\bibitem{song2023learning}
Y.~Song, K.~Shi, R.~Penicka, and D.~Scaramuzza, ``Learning perception-aware agile flight in cluttered environments,'' in \emph{2023 IEEE International Conference on Robotics and Automation (ICRA)}.\hskip 1em plus 0.5em minus 0.4em\relax IEEE, 2023, pp. 1989--1995.

\bibitem{kaufmann2023champion}
E.~Kaufmann, L.~Bauersfeld, A.~Loquercio, M.~M{\"u}ller, V.~Koltun, and D.~Scaramuzza, ``Champion-level drone racing using deep reinforcement learning,'' \emph{Nature}, vol. 620, no. 7976, pp. 982--987, 2023.

\bibitem{ren2022bubble}
Y.~Ren, F.~Zhu, W.~Liu, Z.~Wang, Y.~Lin, F.~Gao, and F.~Zhang, ``Bubble planner: Planning high-speed smooth quadrotor trajectories using receding corridors,'' in \emph{2022 IEEE/RSJ International Conference on Intelligent Robots and Systems (IROS)}.\hskip 1em plus 0.5em minus 0.4em\relax IEEE, 2022, pp. 6332--6339.

\bibitem{zhou2019robust}
B.~Zhou, F.~Gao, L.~Wang, C.~Liu, and S.~Shen, ``Robust and efficient quadrotor trajectory generation for fast autonomous flight,'' \emph{IEEE Robotics and Automation Letters}, vol.~4, no.~4, pp. 3529--3536, 2019.

\bibitem{kamon1997sensory}
I.~Kamon and E.~Rivlin, ``Sensory-based motion planning with global proofs,'' \emph{IEEE transactions on Robotics and Automation}, vol.~13, no.~6, pp. 814--822, 1997.

\bibitem{penicka2022minimum}
R.~Penicka and D.~Scaramuzza, ``Minimum-time quadrotor waypoint flight in cluttered environments,'' \emph{IEEE Robotics and Automation Letters}, vol.~7, no.~2, pp. 5719--5726, 2022.

\bibitem{paull2012sensor}
L.~Paull, S.~Saeedi, M.~Seto, and H.~Li, ``Sensor-driven online coverage planning for autonomous underwater vehicles,'' \emph{IEEE/ASME Transactions on Mechatronics}, vol.~18, no.~6, pp. 1827--1838, 2012.

\bibitem{ye2022efficient}
H.~Ye, N.~Pan, Q.~Wang, C.~Xu, and F.~Gao, ``Efficient sampling-based multirotors kinodynamic planning with fast regional optimization and post refining,'' in \emph{2022 IEEE/RSJ International Conference on Intelligent Robots and Systems (IROS)}.\hskip 1em plus 0.5em minus 0.4em\relax IEEE, 2022, pp. 3356--3363.

\bibitem{sfeir2011improved}
J.~Sfeir, M.~Saad, and H.~Saliah-Hassane, ``An improved artificial potential field approach to real-time mobile robot path planning in an unknown environment,'' in \emph{2011 IEEE international symposium on robotic and sensors environments (ROSE)}.\hskip 1em plus 0.5em minus 0.4em\relax IEEE, 2011, pp. 208--213.

\bibitem{8798720}
F.~Schilling, J.~Lecoeur, F.~Schiano, and D.~Floreano, ``Learning vision-based flight in drone swarms by imitation,'' \emph{IEEE Robotics and Automation Letters}, vol.~4, no.~4, pp. 4523--4530, 2019.

\bibitem{pmlr-v229-liu23a}
\BIBentryALTinterwordspacing
S.~Liu, M.~Xu, P.~Huang, X.~Zhang, Y.~Liu, K.~Oguchi, and D.~Zhao, ``Continual vision-based reinforcement learning with group symmetries,'' in \emph{Proceedings of The 7th Conference on Robot Learning}, ser. Proceedings of Machine Learning Research, J.~Tan, M.~Toussaint, and K.~Darvish, Eds., vol. 229.\hskip 1em plus 0.5em minus 0.4em\relax PMLR, 06--09 Nov 2023, pp. 222--240. [Online]. Available: \url{https://proceedings.mlr.press/v229/liu23a.html}
\BIBentrySTDinterwordspacing

\bibitem{heeg2024learning}
J.~Heeg, Y.~Song, and D.~Scaramuzza, ``Learning quadrotor control from visual features using differentiable simulation,'' \emph{arXiv preprint arXiv:2410.15979}, 2024.

\bibitem{xi2024lightweight}
M.~Xi, H.~Dai, J.~He, W.~Li, J.~Wen, S.~Xiao, and J.~Yang, ``A lightweight reinforcement-learning-based real-time path-planning method for unmanned aerial vehicles,'' \emph{IEEE Internet of Things Journal}, vol.~11, no.~12, pp. 21\,061--21\,071, 2024.

\bibitem{chaplot2020learning}
D.~S. Chaplot, D.~Gandhi, S.~Gupta, A.~Gupta, and R.~Salakhutdinov, ``Learning to explore using active neural slam,'' \emph{arXiv preprint arXiv:2004.05155}, 2020.

\bibitem{xing2024bootstrapping}
J.~Xing, A.~Romero, L.~Bauersfeld, and D.~Scaramuzza, ``Bootstrapping reinforcement learning with imitation for vision-based agile flight,'' \emph{arXiv preprint arXiv:2403.12203}, 2024.

\bibitem{geles2024demonstrating}
I.~Geles, L.~Bauersfeld, A.~Romero, J.~Xing, and D.~Scaramuzza, ``Demonstrating agile flight from pixels without state estimation,'' \emph{Robotics Science and Systems online Proceedings}, no.~20, p. online, 2024.

\bibitem{wen2021mrpb}
J.~Wen, X.~Zhang, Q.~Bi, Z.~Pan, Y.~Feng, J.~Yuan, and Y.~Fang, ``{MRPB 1.0}: A unified benchmark for the evaluation of mobile robot local planning approaches,'' in \emph{2021 IEEE international conference on robotics and automation (ICRA)}.\hskip 1em plus 0.5em minus 0.4em\relax IEEE, 2021, pp. 8238--8244.

\bibitem{heiden2021bench}
E.~Heiden, L.~Palmieri, L.~Bruns, K.~O. Arras, G.~S. Sukhatme, and S.~Koenig, ``{Bench-MR}: A motion planning benchmark for wheeled mobile robots,'' \emph{IEEE Robotics and Automation Letters}, vol.~6, no.~3, pp. 4536--4543, 2021.

\bibitem{toma2021pathbench}
A.-I. Toma, H.-Y. Hsueh, H.~A. Jaafar, R.~Murai, P.~H. Kelly, and S.~Saeedi, ``Pathbench: A benchmarking platform for classical and learned path planning algorithms,'' in \emph{2021 18th Conference on Robots and Vision (CRV)}.\hskip 1em plus 0.5em minus 0.4em\relax IEEE, 2021, pp. 79--86.

\bibitem{xia2020interactive}
F.~Xia, W.~B. Shen, C.~Li, P.~Kasimbeg, M.~E. Tchapmi, A.~Toshev, R.~Mart{\'\i}n-Mart{\'\i}n, and S.~Savarese, ``Interactive gibson benchmark: A benchmark for interactive navigation in cluttered environments,'' \emph{IEEE Robotics and Automation Letters}, vol.~5, no.~2, pp. 713--720, 2020.

\bibitem{moll2015benchmarking}
M.~Moll, I.~A. Sucan, and L.~E. Kavraki, ``Benchmarking motion planning algorithms: An extensible infrastructure for analysis and visualization,'' \emph{IEEE Robotics \& Automation Magazine}, vol.~22, no.~3, pp. 96--102, 2015.

\bibitem{xu2023benchmarking}
Z.~Xu, B.~Liu, X.~Xiao, A.~Nair, and P.~Stone, ``Benchmarking reinforcement learning techniques for autonomous navigation,'' in \emph{2023 IEEE International Conference on Robotics and Automation (ICRA)}.\hskip 1em plus 0.5em minus 0.4em\relax IEEE, 2023, pp. 9224--9230.

\bibitem{rocha2022plannie}
L.~Rocha and K.~Vivaldini, ``Plannie: A benchmark framework for autonomous robots path planning algorithms integrated to simulated and real environments,'' in \emph{2022 International Conference on Unmanned Aircraft Systems (ICUAS)}.\hskip 1em plus 0.5em minus 0.4em\relax IEEE, 2022, pp. 402--411.

\bibitem{shao2024design}
Y.~S. Shao, Y.~Wu, L.~Jarin-Lipschitz, P.~Chaudhari, and V.~Kumar, ``Design and evaluation of motion planners for quadrotors in environments with varying complexities,'' in \emph{2024 IEEE International Conference on Robotics and Automation (ICRA)}.\hskip 1em plus 0.5em minus 0.4em\relax IEEE, 2024, pp. 10\,033--10\,039.

\bibitem{tordesillas2022panther}
J.~Tordesillas and J.~P. How, ``{PANTHER}: Perception-aware trajectory planner in dynamic environments,'' \emph{IEEE Access}, vol.~10, pp. 22\,662--22\,677, 2022.

\bibitem{chen2024apace}
X.~Chen, Y.~Zhang, B.~Zhou, and S.~Shen, ``{APACE}: Agile and perception-aware trajectory generation for quadrotor flights,'' \emph{arXiv preprint arXiv:2403.08365}, 2024.

\bibitem{park2023dlsc}
J.~Park, Y.~Lee, I.~Jang, and H.~J. Kim, ``Dlsc: Distributed multi-agent trajectory planning in maze-like dynamic environments using linear safe corridor,'' \emph{IEEE Transactions on Robotics}, 2023.

\bibitem{song2021autonomous}
Y.~Song, M.~Steinweg, E.~Kaufmann, and D.~Scaramuzza, ``Autonomous drone racing with deep reinforcement learning,'' in \emph{2021 IEEE/RSJ International Conference on Intelligent Robots and Systems (IROS)}.\hskip 1em plus 0.5em minus 0.4em\relax IEEE, 2021, pp. 1205--1212.

\bibitem{zhang2024npe}
Y.~Zhang, C.~Yan, J.~Xiao, and M.~Feroskhan, ``{NPE-DRL}: Enhancing perception constrained obstacle avoidance with nonexpert policy guided reinforcement learning,'' \emph{IEEE Transactions on Artificial Intelligence}, vol.~6, no.~1, pp. 184--198, 2025.

\bibitem{ye2020tgk}
H.~Ye, X.~Zhou, Z.~Wang, C.~Xu, J.~Chu, and F.~Gao, ``{TGK-Planner}: An efficient topology guided kinodynamic planner for autonomous quadrotors,'' \emph{IEEE Robotics and Automation Letters}, vol.~6, no.~2, pp. 494--501, 2021.

\bibitem{zhou2020ego}
X.~Zhou, Z.~Wang, H.~Ye, C.~Xu, and F.~Gao, ``{EGO-Planner}: An esdf-free gradient-based local planner for quadrotors,'' \emph{IEEE Robotics and Automation Letters}, vol.~6, no.~2, pp. 478--485, 2021.

\bibitem{penicka2022learning}
R.~Penicka, Y.~Song, E.~Kaufmann, and D.~Scaramuzza, ``Learning minimum-time flight in cluttered environments,'' \emph{IEEE Robotics and Automation Letters}, vol.~7, no.~3, pp. 7209--7216, 2022.

\bibitem{hirschmuller2007stereo}
H.~Hirschmuller, ``Stereo processing by semiglobal matching and mutual information,'' \emph{IEEE Transactions on pattern analysis and machine intelligence}, vol.~30, no.~2, pp. 328--341, 2007.

\bibitem{falanga2018pampc}
D.~Falanga, P.~Foehn, P.~Lu, and D.~Scaramuzza, ``{PAMPC}: Perception-aware model predictive control for quadrotors,'' in \emph{2018 IEEE/RSJ International Conference on Intelligent Robots and Systems (IROS)}.\hskip 1em plus 0.5em minus 0.4em\relax IEEE, 2018, pp. 1--8.

\bibitem{wang2022geometrically}
Z.~Wang, X.~Zhou, C.~Xu, and F.~Gao, ``Geometrically constrained trajectory optimization for multicopters,'' \emph{IEEE Transactions on Robotics}, vol.~38, no.~5, pp. 3259--3278, 2022.

\bibitem{song2020flightmare}
Y.~Song, S.~Naji, E.~Kaufmann, A.~Loquercio, and D.~Scaramuzza, ``Flightmare: A flexible quadrotor simulator,'' in \emph{Proceedings of the 2020 Conference on Robot Learning}, 2021, pp. 1147--1157.

\bibitem{gazebo}
N.~Koenig and A.~Howard, ``Design and use paradigms for {Gazebo}, an open-source multi-robot simulator,'' in \emph{2004 IEEE/RSJ International Conference on Intelligent Robots and Systems (IROS) (IEEE Cat. No.04CH37566)}, vol.~3, 2004, pp. 2149--2154 vol.3.

\bibitem{quigley2009ros}
M.~Quigley, K.~Conley, B.~Gerkey, J.~Faust, T.~Foote, J.~Leibs, R.~Wheeler, A.~Y. Ng \emph{et~al.}, ``{ROS}: an open-source robot operating system,'' in \emph{ICRA workshop on open source software}, vol.~3, no. 3.2.\hskip 1em plus 0.5em minus 0.4em\relax Kobe, Japan, 2009, p.~5.

\end{thebibliography}
